\documentclass[11pt]{article}

\usepackage[final]{acl}

\usepackage{times}
\usepackage{latexsym}

\usepackage[T1]{fontenc}

\usepackage[utf8]{inputenc}

\usepackage{microtype}

\usepackage{inconsolata}

\usepackage{graphicx}
\usepackage{subcaption}

\usepackage{tabularx}
\usepackage{fontspec}
\usepackage{xeCJK}
\usepackage{booktabs}
\usepackage{amsmath}
\usepackage{amssymb}

\usepackage{color, soul}

\title{Rethinking Metrics for Lexical Semantic Change Detection}

\author{Roksana Goworek \\
  Queen Mary University of London \\
  \texttt{r.goworek@qmul.ac.uk} \\\And
  Haim Dubossarsky \\
  Queen Mary University of London \\
  The Alan Turing Institute \\
  University of Cambridge \\
  \texttt{h.dubossarsky@qmul.ac.uk} \\}

\begin{document}
\maketitle
\begin{abstract}

Lexical semantic change detection (LSCD) increasingly relies on contextualised language model embeddings, yet most approaches still quantify change using a small set of semantic change metrics, primarily Average Pairwise Distance (APD) and cosine distance over word prototypes (PRT). We introduce Average Minimum Distance (AMD) and Symmetric Average Minimum Distance (SAMD), new measures that quantify semantic change via local correspondence between word usages across time periods. Across multiple languages, encoder models, and representation spaces, we show that AMD often provides more robust performance, particularly under dimensionality reduction and with non-specialised encoders, while SAMD excels with specialised encoders. We suggest that LSCD may benefit from considering alternative semantic change metrics beyond APD and PRT, with AMD offering a robust option for contextualised embedding-based analysis.
\end{abstract}

\begin{figure}[t]
    \begin{subfigure}[t]{0.48\linewidth}
        \centering
        \includegraphics[width=\linewidth]{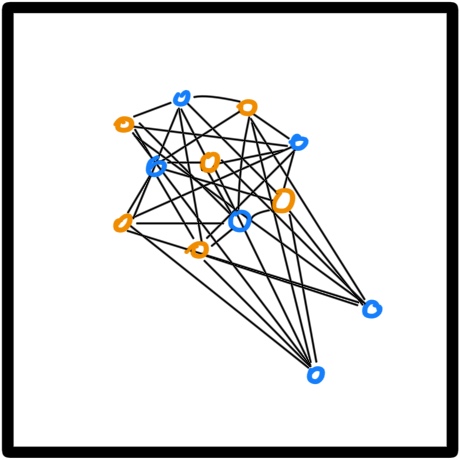}
        \caption{APD: average distance over all cross-corpus usage pairs.}
        \label{fig:amd_intuition_apd}
    \end{subfigure}\hfill
    \begin{subfigure}[t]{0.48\linewidth}
        \centering
        \includegraphics[width=\linewidth]{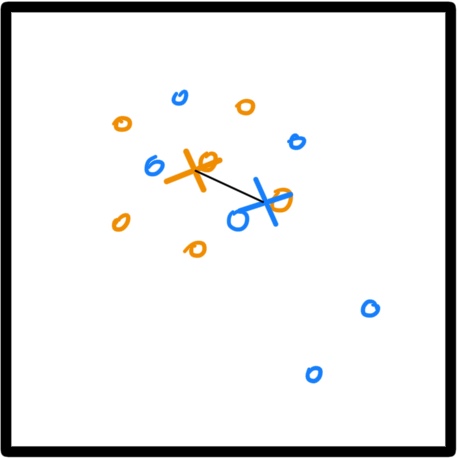}
        \caption{PRT: cosine distance between corpus centroids.}
        \label{fig:amd_intuition_prt}
    \end{subfigure}

    \vspace{0.5em}

    \begin{subfigure}[t]{0.48\linewidth}
        \centering
        \includegraphics[width=\linewidth]{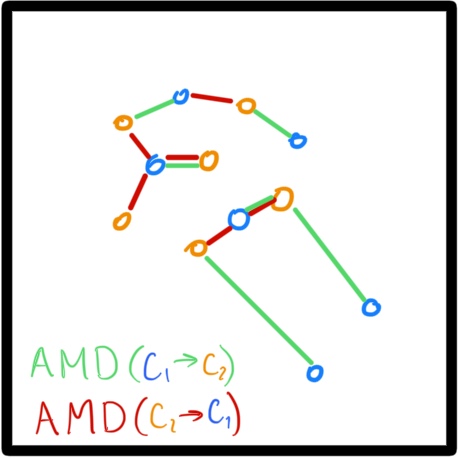}
        \caption{AMD: nearest cross-corpus links ($C_1\!\rightarrow\!C_2$ in green, $C_2\!\rightarrow\!C_1$ in red).}
        \label{fig:amd_intuition_amd}
    \end{subfigure}\hfill
    \begin{subfigure}[t]{0.48\linewidth}
        \centering
        \includegraphics[width=\linewidth]{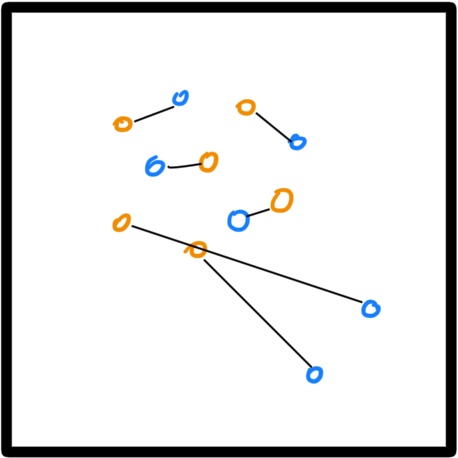}
        \caption{SAMD: greedy one-to-one matching between cross-corpus usages.}
        \label{fig:amd_intuition_samd}
    \end{subfigure}

    \caption{Illustrative schematic contrasting LSCD distance measures on usage embeddings from two corpora (blue: $C_1$, yellow: $C_2$).}
    \label{fig:amd_intuition_2x2}
\end{figure}

\section{Introduction}

Lexical Semantic Change Detection (LSCD) aims to automatically quantify how word meanings evolve over time through the analysis of diachronic text corpora \cite{tahmasebi2023computational}.
Recent shared tasks and benchmarks have established a standard evaluation setting in which systems must determine, for a given target word and two time periods, whether its meaning has changed and to what extent \cite{schlechtweg2020unsupervised, sutter2024lscdbenchmark}.
With the rise of contextualised language models, LSCD approaches have increasingly moved from static type-level representations to usage-level embeddings, in which each occurrence of a target word is encoded as a high-dimensional vector reflecting its local context \cite{martinc2020leveraging}.
Semantic change is then quantified by comparing the distributions of these usage embeddings across corpora, most commonly using semantic change metrics such as Average Pairwise Distance (APD) or distances between corpus-specific word prototypes (PRT) \cite{kutuzov2020uio}.

While these approaches achieve strong performance on benchmark datasets, the choice of semantic change metric implicitly encodes assumptions about what constitutes semantic change.
Metrics such as APD and PRT aggregate information globally—either across all cross-corpus usage pairs or via a single centroid per corpus—which makes them well suited to capturing broad distributional shifts.
This global aggregation can obscure localised phenomena, such as the emergence of a new sense, the disappearance of an older one, or changes affecting only a subset of usages whose occurrences might be too rare to notice.

In this paper, we introduce \textbf{Average Minimum Distance (AMD)} and its \textbf{Symmetric} variant \textbf{SAMD}, a pair of semantic change metrics based on usage-level nearest cross-corpus correspondence.
For each usage in one corpus, AMD measures the distance to its closest counterpart in the other corpus and averages these minimum distances in both directions, while SAMD further enforces a one-to-one alignment between usage instances across periods.
This formulation is grounded in the intuition that semantic change is reflected in how well individual usages from one corpus can be matched to semantically similar usages in another, rather than solely in how far two usage distributions are on average.
As illustrated in \autoref{fig:amd_intuition_2x2}, AMD and SAMD emphasise local correspondence between usages and naturally capture usage-level changes.

AMD permits a directional decomposition, distinguishing how well earlier-corpus usages are explained by later-corpus usages and vice versa.
This makes it possible to analyse asymmetric changes such as broadening/narrowing or sense emergence and disappearance.
Moreover, because AMD and SAMD rely on usage-level correspondences rather than global distributional geometry, they can be applied more meaningfully in reduced or restructured representation spaces.

Our experiments show that AMD and SAMD provide robust and complementary alternatives to existing semantic change metrics, particularly when embeddings are projected into lower-dimensional spaces, or when non-specialised encoders are used. We release code, results, and generated definitions to reproduce our experiments and construct definition-based spaces.\footnote{\url{https://github.com/roksanagow/Rethinking_LSCD_Metrics}}


\section{Related Work}

\subsection{Semantic change metrics for LSCD}

Early work on LSCD was developed in the context of static word embeddings, where a word is represented by a single vector per corpus \cite{kim-etal-2014-temporal, hamilton-etal-2016-cultural, schlechtweg19_acl}. In this setting, semantic change is naturally quantified by comparing word representations across corpora. With the introduction of contextualised models this paradigm was reintroduced using PRT, which averages the representations of all usages of a particular word, leaving a single vector for subsequent analysis as in static models.

In contrast, global-distribution-based metrics such as Average Pairwise Distance (APD) are motivated by the availability of contextualised usage embeddings and explicitly compare distributions of individual word occurrences across corpora \cite{kutuzov-giulianelli-2020-semeval,kutuzov2022lessons}. By exploiting the richer structure of contextualised representations, APD extends prototype-based approaches and has become a standard usage-level semantic change metric in contemporary LSCD benchmarks, usually at the top of the leader-board \cite{periti2024systematic}.

\subsection{Robustness and discovery-oriented LSCD}

Recent work has raised concerns about whether strong benchmark performance translates into reliable semantic-change discovery in realistic settings \cite{hamilton2016diachronic, kurtyigit2021lexical}.
In particular, \citet{umarova-etal-2025-current} show that when LSCD systems are applied to full vocabularies, many high-scoring words correspond to topical or domain variation rather than genuine semantic change.
This finding highlights a gap between curated benchmark evaluations and large-scale discovery scenarios, and suggests that semantic change metrics emphasising global distributional shifts may be vulnerable to confounds when applied broadly \cite{schlechtweg19_acl,dubossarsky2019timeout}.

\subsection{Interpretable LSCD}

In parallel, several strands of research have explored more interpretable frameworks for LSCD.
Hypothesis-driven approaches incorporate expert-defined sense distinctions to track fine-grained semantic change, arguing that unsupervised methods may fail to align representations with linguistically meaningful senses \cite{cassotti-etal-2023-xl}.
\citet{fedorova2024definitions} propose generating LLM-based definitions for individual usages and detecting change by comparing distributions of induced sense labels, while SCDTour \cite{aida2025scdtour} identifies interpretable axes of change in static embedding spaces.
These approaches demonstrate the potential of interpretable representations, but rely on specialised pipelines or additional modelling assumptions, and typically retain global semantic change metrics such as APD or PRT for quantification.

\section{Method}
\label{sec:method}

\subsection{Task Formulation and Notation}
\label{subsec:task}

We follow the standard lexical semantic change detection (LSCD) setting.
For a given language, we are given two diachronic corpora
\(C_1\) and \(C_2\), corresponding to an earlier and a later
time period.
Let \(\mathcal{W}\) denote the set of target lemmas for this language.
For each \(w \in \mathcal{W}\) and each corpus \(C_t\) (\(t \in \{1,2\}\)),
we extract all sentences (or sentence-like contexts) that contain
an occurrence of \(w\).
We write
\[
U_t(w) = \{ u^{(t)}_1, \dots, u^{(t)}_{n_t} \}
\]
for the set of usage instances of \(w\) in period \(t\),
where each \(u^{(t)}_i\) is a sequence of tokens with a marked
target position.

\noindent
Let \(f_\theta\) be a contextual encoder that maps a usage \(u\)
with a marked target token to a contextual embedding
\(v = f_\theta(u) \in \mathbb{R}^D\).
We denote the resulting sets of usage embeddings as
\[
V_t(w) = \{ v^{(t)}_1, \dots, v^{(t)}_{n_t} \}
\subset \mathbb{R}^D.
\]

For notational simplicity, in the remainder of this section we write
\(A = V_1(w)\) and \(B = V_2(w)\) for the sets of usage embeddings
of a target word \(w\) in the two time periods.
Given \(A\) and \(B\), we compute a graded semantic change score
using the LSCD metrics described in \autoref{subsec:metrics}.

\noindent\textbf{Evaluation.} For evaluation, each target word is assigned a single semantic
change score by a given metric, and performance is measured as the
Spearman rank correlation between these scores and the gold graded
change scores provided with the datasets, averaged across all
target words.

Our method is encoder-agnostic: it operates on any token-level
embedding model \(f_\theta\), and we compare it across multiple encoders described in \autoref{subsec:encoders}.

\noindent\textbf{Datasets.}
We evaluate on LSCD benchmarks in seven languages: English, German, Swedish, and Latin datasets from SemEval-2020 Task 1 \cite{schlechtweg2020semeval}, the Spanish semantic change dataset from LSCDiscovery \cite{zamora-reina2022lscdiscovery}, the Norwegian dataset NorDiaChange \cite{kutuzov2022nordiachange}, and the Chinese semantic shift benchmark \cite{chen-etal-2022-lexicon} for Chinese. Each benchmark provides diachronic corpus pairs with human-annotated semantic change scores.

\subsection{LSCD Metrics}
\label{subsec:metrics}

All metrics use cosine distance, defined as
\(\delta(x,y) = 1 - \cos(x,y)\).

\paragraph{Average Pairwise Distance (APD)} computes the mean distance between
all cross-corpus usage pairs:
\[
\mathrm{APD}(A,B)
= \frac{1}{|A||B|}
  \sum_{a \in A} \sum_{b \in B} \delta(a,b).
\]
APD captures global divergence between usage distributions but is
dominated by the bulk of points, making it less sensitive to small
or emerging usage clusters.

\paragraph{Prototype Distance (PRT)} compares the centroids of the two periods.
Let

\(
\mu(A) = \frac{1}{|A|}\sum_{a \in A} a
\).

Then
\[
\mathrm{PRT}(A,B)
= \delta\big( \mu(A), \mu(B) \big).
\]
PRT is computationally efficient but reduces the representations of a word in each period to a
single point, potentially obscuring sense-specific changes.


\paragraph{Average Minimum Distance (AMD).}
We introduce AMD, a nearest-neighbour–based
metric designed to capture local discrepancies between periods.
For a point \(x\) and a set \(Y\), we define
\[
\mathrm{nndist}(x,Y) = \min_{y \in Y} \delta(x,y).
\]
The directional score is
\[
\mathrm{AMD}(A \rightarrow B)
= \frac{1}{|A|} \sum_{a \in A} \mathrm{nndist}(a,B),
\]
with \(\mathrm{AMD}(B \rightarrow A)\) defined analogously.
The symmetric score is
\[
\mathrm{AMD}(A,B)
= \tfrac{
\mathrm{AMD}(A \rightarrow B)
+ \mathrm{AMD}(B \rightarrow A)}{2}.
\]

\paragraph{Symmetric Average Minimum Distance (SAMD).}
While AMD allows many-to-one nearest-neighbour mappings, we further introduce
SAMD, which enforces one-to-one correspondence between usage instances across periods.
We sample an equal number of usages from both corpora, corresponding to the minimum available across the two periods, such that $|A| = |B|$.
Define the pairwise distances \(D_{ij}=\delta(a_i,b_j)\).
SAMD greedily selects the smallest remaining distance, removes the corresponding
row and column, and repeats until all usage pairs are matched.\footnote{We also evaluated optimal one-to-one matching via the Hungarian algorithm and found results to be near-identical to the greedy algorithm used here.} The final score is
\[
\mathrm{SAMD}(A,B)
= \frac{1}{|A|} \sum_{k=1}^{|A|} \delta(a_{i_k}, b_{j_k}).
\]





AMD captures local shifts by matching each usage to its closest counterpart,
but may map multiple points onto the same neighbour. 
This behaviour can be problematic in highly hub-dominated spaces \cite{radovanovic_hubness, Dinu2014ImprovingZL} (see \autoref{app:hubness} for an analysis of hubness in the selected embedding spaces). This phenomenon is of particular concern in LSCD, where a small number of modern-style 
usages in the historical corpus may emerge as nearest neighbours for a 
disproportionate number of modern usages, despite lacking semantic 
correspondence.
SAMD mitigates this effect by approximating a minimal one-to-one alignment
between periods, yielding a metric that is sensitive to distributional shifts as well as cross-corpus correspondence.

\subsection{Models}
\label{subsec:encoders}

We evaluate AMD across a diverse set of pretrained language models in order to assess its robustness.
We include \textbf{XL-LEXEME} \cite{cassotti-etal-2023-xl}, a multilingual encoder, based on XLM-RoBERTa, explicitly fine-tuned for word-in-context (WiC) discrimination. XL-LEXEME has been shown to achieve state-of-the-art performance on LSCD.

\paragraph{Multilingual encoders.}
We compare several widely used multilingual encoders without any fine-tuning, including XLM-RoBERTa \cite{conneau-etal-2020-unsupervised}, a strong general-purpose multilingual baseline; the recently released mmBERT \cite{marone2025mmbert}, designed for improved multilingual semantic alignment; RemBERT \cite{rembert}, which emphasises language-balanced pretraining across many languages; and multilingual-E5 \cite{wang2024multilingual}, a retrieval-oriented model optimised for representing semantic similarity. These models are commonly employed in cross-lingual and multilingual semantic tasks, and allow us to test whether different LSCD metrics remain effective when applied to generic contextual representations.

\paragraph{Monolingual encoders.}
We also evaluate various monolingual encoders, selecting one model per language: RoBERTa for English \cite{roberta}, GBERT for German \cite{chan-etal-2020-germans}, BammanBurns-BERT for Latin \cite{bamman2020latin}, Spanish BERT \cite{CaneteCFP2020}, Chinese RoBERTa \cite{cui-etal-2020-revisiting}, NB-BERT for Norwegian \cite{kummervold-etal-2021-operationalizing}, and Megatron-BERT for Swedish \cite{malmsten2020playing}. These models are trained on large language-specific corpora and provide strong language-specific contextual representations for LSCD. Focusing on monolingual encoders is meant to test AMD and SAMD under conditions mimicking low resource (and historical) languages (e.g., Latin, Ancient Greek), or on specialized domains (e.g., medical, legal), where multilingual models struggle or may not be available.

\subsection{Reduced Space Construction}
\label{subsec:def-space}

To analyse the behaviour of LSCD metrics under varying
representational constraints, we evaluate them not only on the
original contextual embedding space, but also on several
reduced-dimensional spaces constructed from the same usage
embeddings. These spaces preserve different aspects of the
original representations: (i) full embeddings, which retain all
information; (ii) a definition-based space, which preserves
semantically interpretable information; (iii) principal components,
which preserve maximal global variance; and (iv) randomly selected
dimensions, which preserve neither semantic structure nor global
variance in any principled way.

All LSCD metrics are applied either to the original sets
\(A,B \subset \mathbb{R}^D\) or to their transformed counterparts
\(\phi(A), \phi(B) \subset \mathbb{R}^K\), depending on the
representation space.

\paragraph{Definition-space.}
For each target word \(w\), we generate a finite set of textual
definitions
\[
\mathcal{D}(w) = \{ d_1, \dots, d_K \}
\]
using \textbf{Gemini~2.5~Pro} \cite{comanici2025gemini}, a multilingual large language model.
The same prompt, translated into each target language, is used
across languages (see \autoref{tab:definition_prompts} for prompts and \autoref{tab:num_defs} for the amounts of generated definitions). 
We do not sample definitions from existing dictionaries, as it is difficult to obtain resources that provide multiple definitions consistently for all target words. 
Moreover, dictionary entries for different words are often authored by different lexicographers, which can introduce variation in sense granularity and segmentation across entries. 
Our setup instead mirrors a scenario in which a single lexicographer provides a set of candidate definitions per investigated word.

Each definition is prepended with the target word, yielding inputs
of the form \texttt{$w$: $d_k$} where \texttt{$d_k$} is the $k^{th}$ definition for word \texttt{$w$}.
We obtain contextualised embeddings for the target word within each
definition using the same encoder \(f_\theta\) used to encode the usages, resulting in
definition embeddings
\[
z_k = f_\theta(\texttt{$w$: $d_k$}) \in \mathbb{R}^D.
\]

Given a usage embedding \(v \in A \cup B\), we project it into
definition-space by computing its cosine distance to each
definition embedding:
\[
\phi_{\text{def}}(v) =
\big(
\delta(v, z_1), \dots, \delta(v, z_K)
\big)
\in \mathbb{R}^K.
\]
The resulting vectors are treated as standard embeddings in
\(\mathbb{R}^K\), to which APD, PRT, AMD and SAMD are applied directly.

\paragraph{Principal component analysis (PCA).}
As a variance-preserving but non-interpretable projections, we apply PCA independently for each target word, using all its usage embeddings across both periods. Each embedding \(v \in \mathbb{R}^D\) is projected onto the top \(K\) principal components. Unless stated otherwise, \(K\) matches the number of definitions used in the definition-space projection for direct comparison, while the stress test varies \(K\) to assess sensitivity to number of dimensions.

\paragraph{Random dimension selection.}
An extreme case of dimensionality is introduced by randomly selecting \(K\) embedding dimensions independently for each target word. The same dimensions are retained for all usages of that word across both periods, allowing us to isolate the effect of dimensionality reduction from that of principled subspace construction. Like for PCA, K matches the number of definitions generated for a given word to facilitate comparison, while the stress test varies K to assess sensitivity of metrics to number of dimensions.

\section{Results}
\label{sec:results}

We evaluate LSCD metrics across representation spaces, encoder families, and languages, with a focus on their robustness under varying representational constraints. All results report Spearman correlations with graded semantic change scores and are averaged across languages and encoder models unless stated otherwise.

\begin{figure}[h]
    \centering
    \includegraphics[width=\linewidth]{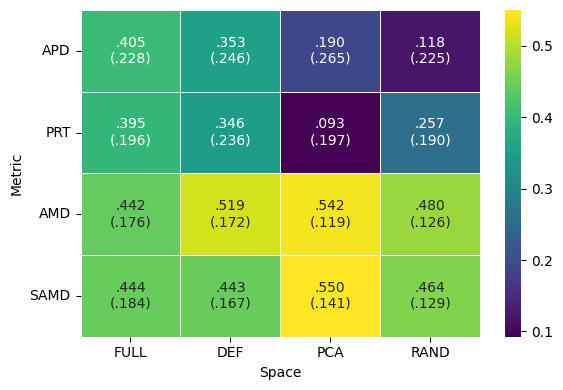}
    \caption{Performance (Spearman correlations) across metrics and spaces, averaged over languages and encoders. Std in brackets.}
    \label{fig:metric_space}
\end{figure}

\paragraph{Overall comparison.}
\autoref{fig:metric_space} presents a high-level comparison of APD, PRT, AMD, and SAMD across full embeddings and reduced-dimensional spaces.
The results show a consistent pattern: correspondence-based metrics have the best performance (in terms of average Spearman correlations) and the most stable (in terms of lowest variance) across all four types of embedding spaces (FULL, DEF, PCA, RAND). 

The best performance of AMD and SAMD is when they are paired with PCA dimensionality reduction, suggesting that PCA may reduce irrelevant noise in the embeddings. Notably, AMD and SAMD can uniquely withstand the noise elicited by the more aggressive dimensionality reduction methods (PCA and RAND), which can even improve performance.

Moreover, AMD is the only metric that significantly gains from projecting its representations to the definition space, while APD and PRT show a significant drop in performance. 
In the interpretable definition-based space, AMD achieves the highest average performance, suggesting that, in this space, semantic change is better captured through differences in the alignment of the overall shape of usage embedding distributions, rather than through shifts in their distribution concentration.
In contrast, AMD and SAMD perform similarly in FULL, PCA and RAND spaces, while remaining markedly more robust than APD and PRT.

Overall, these results demonstrate that local correspondence metrics provide both stronger and more stable estimates of semantic change across diverse representation spaces.

\begin{figure}[h]
    \centering
    \includegraphics[width=\linewidth]{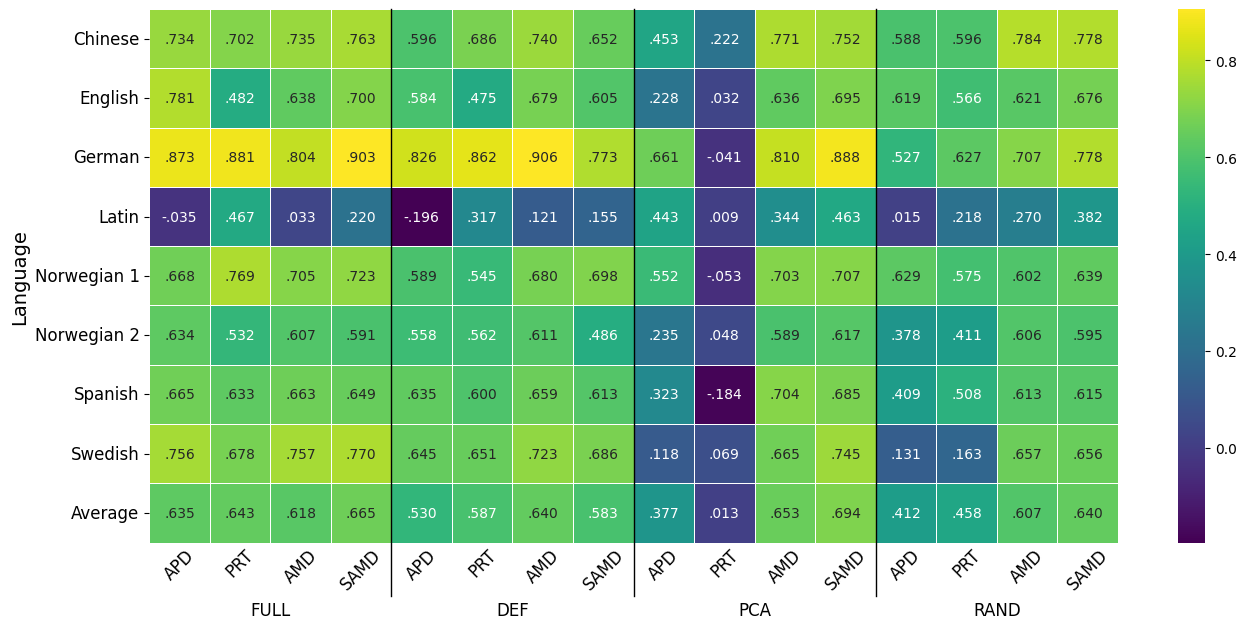}
    \caption{Performance of APD, PRT, AMD and SAMD using XL-LEXEME embeddings across representation spaces. Each row corresponds to a language; columns show metric–space combinations.}
    \label{fig:xll_lang_ms}
\end{figure}

\paragraph{Specialised encoder: XL-LEXEME.}
We next examine results obtained with XL-LEXEME, a model explicitly fine-tuned for LSCD (\autoref{fig:xll_lang_ms}).
Here too the best performing approach is SAMD with PCA, achieving an average Spearman correlation of .694 across all languages, surpassing APD (.635) and PRT (.643) on full embeddings. 
SAMD also outperforms standard metrics in the full embedding space. 

In the definition-based space, AMD yields the highest correlation (.640), again highlighting the effectiveness of directional nearest-neighbour alignment in structured semantic spaces. However, the gains from combining AMD with DEF space are markedly larger for non-specialised encoders than for XL-LEXEME, likely because XL-LEXEME’s full embedding space is already organised in a sense-aware fashion through its fine-tuning.

As observed in the aggregate analysis in \autoref{fig:metric_space}, both AMD and SAMD remain robust under PCA and random dimensionality reduction, whereas APD and PRT exhibit detrimental degradation.
These findings indicate that even for specialised encoders, local correspondence metrics provide robust estimates of semantic change, especially under dimensionality reduction.

\begin{figure*}[!h]
\centering
\begin{subfigure}{0.5\textwidth}
    \includegraphics[width=\linewidth]{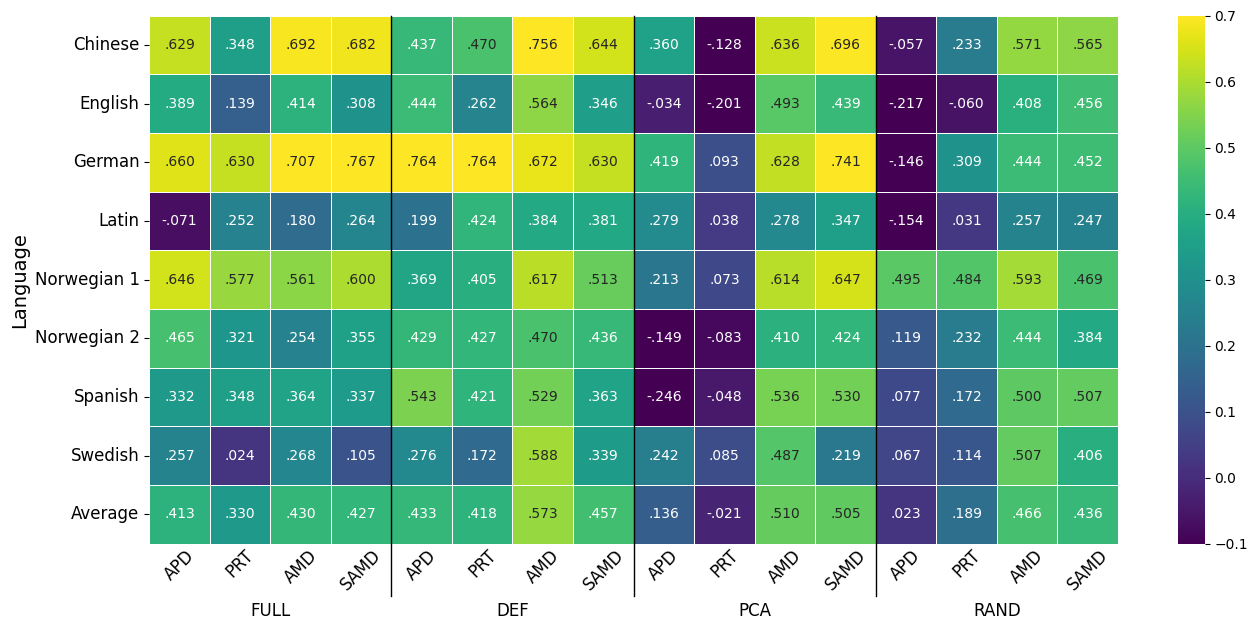}
    \caption{Monolingual encoders (one per language).}
    \label{fig:mono_lang_ms}
\end{subfigure}\hfill
\begin{subfigure}{0.5\textwidth}
    \includegraphics[width=\linewidth]{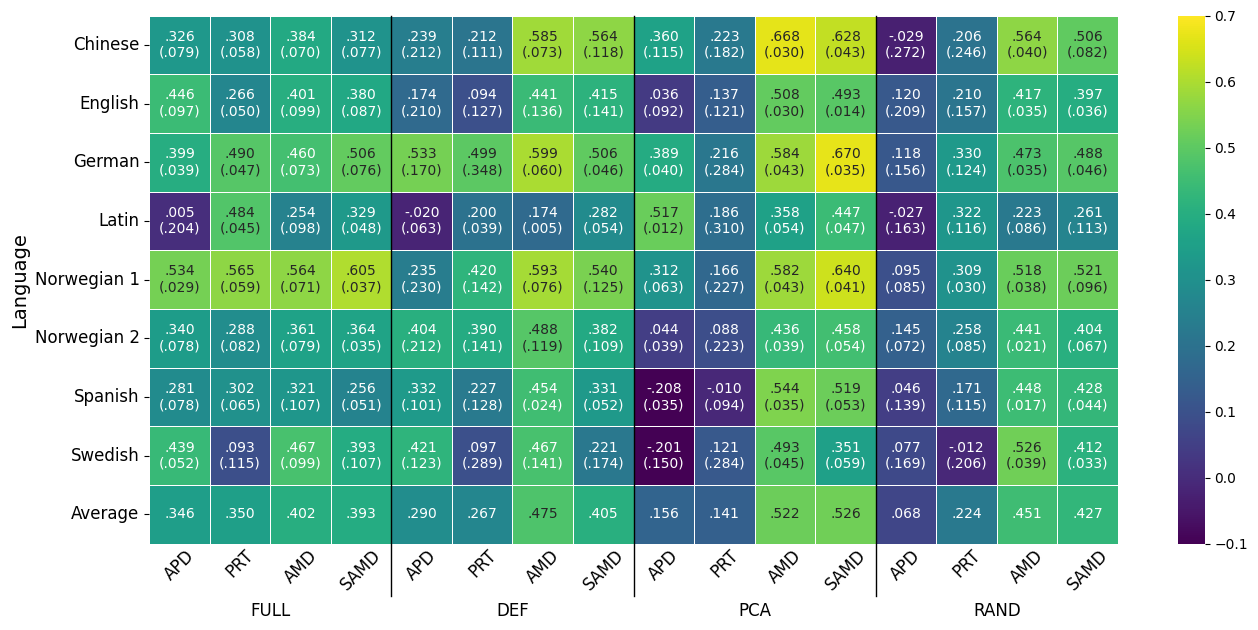}
    \caption{Multilingual encoders (averaged over models; std).}
    \label{fig:multi_lang_ms}
\end{subfigure}
\caption{Performance of APD, PRT, AMD and SAMD for non-specialised encoders.}
\label{fig:non_specialised}
\end{figure*}

\paragraph{Non-specialised encoders.}
Figures \ref{fig:mono_lang_ms} and~\ref{fig:multi_lang_ms} summarise results for language-specific monolingual encoders and general-purpose multilingual encoders, respectively.
Across both settings, performance is lower than for specialised encoders. Nevertheless, the benefit of our two novel metrics is clear. 
Notably, monolingual encoders consistently outperform the averaged multilingual models, suggesting that language-specific models may be better suited to capturing fine-grained usage distinctions. This aligns with prior findings questioning the effectiveness of multilingual representations, or multilingual fine-tuning for sense-level tasks \cite{goworek-dubossarsky-2025-multilinguality, goworek-etal-2025-senwich}.
Fine-tuning such monolingual encoders on WiC may therefore yield stronger performance than relying on a single general-purpose multilingual model.

For monolingual encoders (\autoref{fig:mono_lang_ms}), AMD paired with the definition-based space yields the strongest average performance, substantially outperforming all other metric–space combinations. This could suggest that the sense-aware organisation of the embedding space is conducive to highlighting change which manifests as difficult-to-pair usages across corpora.
The next best results are achieved by AMD and SAMD under PCA reduction, supporting previous results.
On full embeddings, the correspondence-based metrics again outperform APD and PRT, and these global-shift metrics collapse under PCA and random reduction, often approaching zero correlation.

A similar pattern emerges for multilingual encoders (\autoref{fig:multi_lang_ms}).
The best average performance is obtained by SAMD and AMD under PCA reduction (.526 and .522), followed by AMD in the definition space (.475). One possible explanation for the weaker benefit of the AMD–DEF combination compared to monolingual models is that multilingual encoders may exhibit weaker sense-level organisation within each language, as they must represent a much broader multilingual vocabulary, making semantic projections less effective.

In the full space, AMD and SAMD remain around .400, while APD and PRT achieve a Spearman correlation of approximately .350.
Again, APD and PRT exhibit severe sensitivity to dimensionality reduction.

Taken together, these results confirm that AMD and SAMD provide substantially more robust semantic change estimates than global metrics, particularly in constrained or lower-dimensional representation spaces.

\begin{figure}[h]
    \centering
    \includegraphics[width=\linewidth]{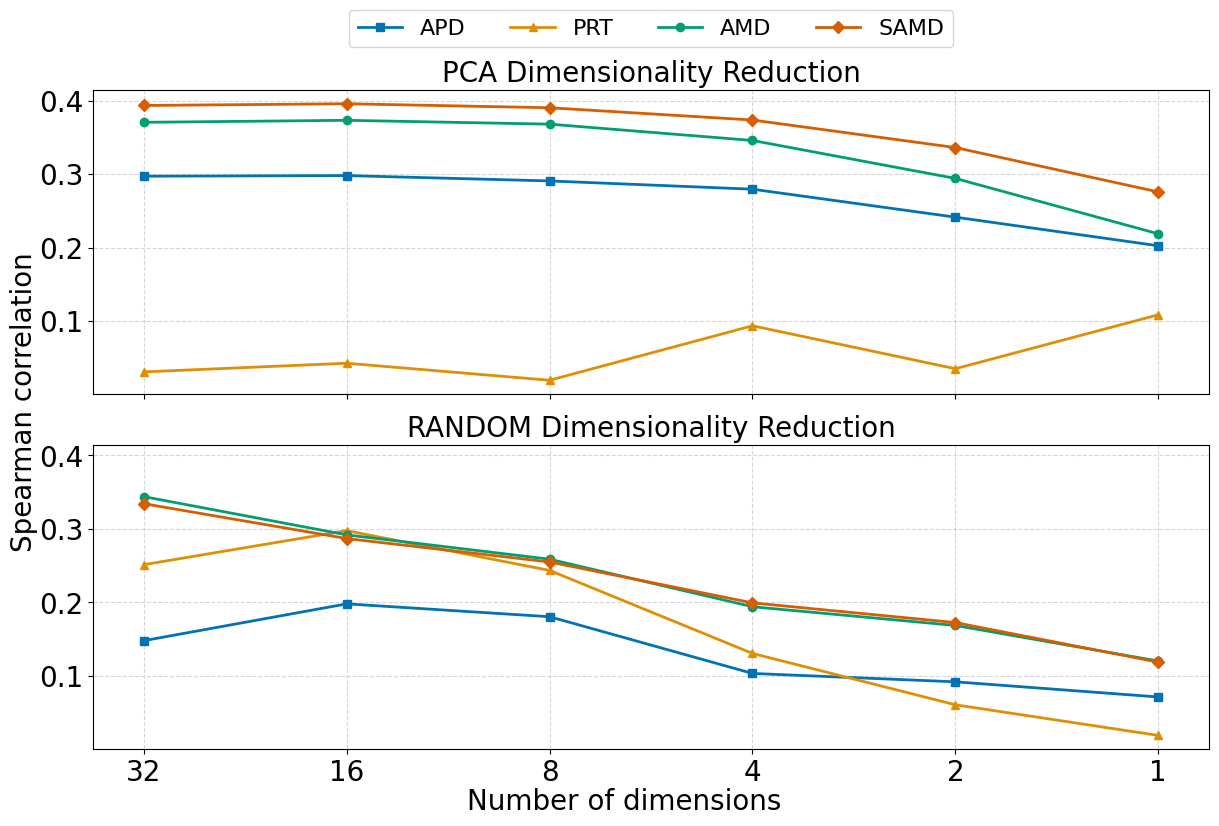}
    \caption{Stress test of metrics under progressive dimensionality reduction. 
    Spearman correlations are averaged across languages and encoders as embedding dimensionality is reduced by factors of two using PCA or random dimension selection.}
    \label{fig:stress_test}
\end{figure}

\subsection{Stress Test}
To further analyse robustness, we progressively reduce embedding dimensionality by factors of two using PCA and random dimension selection (\autoref{fig:stress_test}).
PRT fails almost completely under PCA, yielding correlations close to zero even at moderate dimensionalities.
APD yields average correlations around .300 in low-dimensional spaces.
In contrast, AMD and SAMD exhibit greater robustness, maintaining correlations near .400 up until only 4 dimensions are retained.
Under PCA, SAMD shows slightly higher robustness than AMD, while under random reduction both metrics perform almost identically.
Across most numbers of dimensions, APD and PRT remain consistently worse by almost .100.
These results highlight that global distribution-based metrics rely heavily on high-dimensional embedding structure, whereas local correspondence metrics retain meaningful performance under dimensionality reduction.

\section{Discussion}
\label{sec:discussion}


Our results show that while some metrics are, on average, better than others, their effectiveness largely depends on encoder choices (e.g., multi or monolingual, finetuned or not) as well as on the embedding quality and representational constraints (e.g., dimensionality reduction or semantic projection). 
This aligns with recent work showing that LSCD performance is highly sensitive to modelling choices and evaluation conditions, with no single method consistently dominating across settings or languages \cite{periti2024systematic}. 

This is also supported by work showing that LSC itself may manifest along different dimensions of change such as affective or pragmatic shifts, and that no single model is found to be the best across different change types \cite{baes-etal-2025-lsc, goworek-dubossarsky-2024-toward}.

\paragraph{Local correspondence vs.\ global aggregation.}
Prototype-based measures such as PRT and global usage-level metrics such as APD summarise semantic change by aggregating information across all usages.
As a result, they are most sensitive to changes that affect the overall distribution of usages or their centroid, making them effective when semantic change manifests as a broad redistribution of usage embeddings. 
This behaviour has been noted in prior work, where APD and PRT were shown to exhibit dataset-dependent performance, with preferences for one over the other correlating with properties of the target-word change score distributions rather than a single metric consistently dominating across conditions \cite{kutuzov2022lessons, martinc2020leveraging}.

In contrast, the focus of AMD and SAMD on local correspondence between usages makes them particularly sensitive to changes affecting only a subset of usages
, which may be diluted by global aggregation.
While AMD allows multiple usages to map to the same nearest neighbour and can be decomposed into more interpretable directional components, SAMD enforces a one-to-one alignment between periods, making it sensitive not only to changes in distributional shape but also to shifts in the concentration of usage embeddings.
These observations suggest that semantic change metrics are not interchangeable. Their behaviour differs systematically depending on how they aggregate information across usages and how they interact with representational constraints \cite{kutuzov2022lessons}.

\paragraph{Robustness and dimensions.}
Our results highlight that semantic change metrics differ substantially in their sensitivity to representational quality and embedding structure. 
Global distribution-based metrics such as APD and PRT degrade sharply under dimensionality reduction and are less optimal with less specialised embedding spaces, whereas local correspondence-based metrics such as AMD and SAMD remain considerably more robust.

This finding resonates with earlier observations that LSCD performance is strongly influenced by embedding alignment quality and representational noise \cite{martinc2020leveraging, giulianelli-etal-2020-analysing}. 
It also complements recent work questioning the benefits of multilingual fine-tuning and representations for fine-grained semantic tasks, where monolingual model often outperform general-purpose multilingual models, even after fine-tuning \cite{goworek-dubossarsky-2025-multilinguality, goworek-etal-2025-senwich}.

Our findings may additionally connect to concerns about large-scale semantic-change discovery \cite{umarova-etal-2025-current}. 
Methods optimised for curated benchmark evaluations frequently prioritise detecting broad semantic shift over more realistic and nuanced semantic change when applied across full vocabularies.
By focusing on usage-level correspondence and exhibiting greater robustness to less specialised or manipulated representations, AMD and SAMD may be a useful choice in exploratory settings.. However, like all embedding-based approaches, their effectiveness depends on the availability of sufficiently representative corpora that capture the full range of a word’s usages.


\paragraph{Interpretability and lower-dimensional representations.}
Low-dimensional representations such as definition-based spaces make it possible to relate model behaviour to human-understandable semantic dimensions, enabling targeted qualitative analysis of how and why meanings change.
In our experiments, AMD and SAMD remain competitive under such representations and often outperform APD applied to full embeddings.
AMD appears particularly suitable for deeper analysis due to its directional decomposition, which enables fine-grained inspection of asymmetric mismatches between periods via its directional components.
In contrast, while SAMD yields strong quantitative performance, its one-to-one matching formulation does not directly support directional interpretability.

These findings align with efforts to move beyond black-box approaches to semantic change detection. Interpretable semantic representations that combine corpus-derived embeddings with structured lexical evidence can provide richer frameworks for understanding semantic change, as explored in linked-data approaches for multilingual diachronic analysis \cite{armaselu2024multilingual}.
Other examples include creating interpretable transformations of embedding spaces into structured semantic dimensions, facilitating both analysis and robustness \cite{cassotti-etal-2024-using, aida2025scdtour}. 
Unlike methods that induce latent sense labels or rely on static prototypes, definition-based projections offer fixed, usage-independent semantic dimensions that can be paired with usage-level metrics to support transparent analysis.
More broadly, semantic change metrics that remain robust under dimensionality reduction enable LSCD approaches that balance interpretability with quantitative performance.

\paragraph{Outlook.}
Overall, our results argue against a one-size-fits-all approach to semantic change metrics. 
This conclusion aligns with recent large-scale evaluations demonstrating that LSCD methods exhibit highly variable performance depending on the type of semantic change, representation quality, and evaluation conditions \cite{baes-etal-2025-lsc, periti2024systematic}.

Prototype-based measures, global usage-level metrics, and local correspondence-based metrics may therefore behave differently across research settings, such as low-resource languages, corpora from varying time periods, or specialised domains where embedding quality is weaker.
AMD and SAMD should be viewed not as replacements for existing measures, but as complementary additions that expand the methodological toolkit for contextualised embedding-based LSCD.

\section{Conclusion}
\label{sec:conclusion}

In this paper, we introduced Average Minimum Distance (AMD) and its Symmetric variant, SAMD, two usage-level metrics for lexical semantic change detection. These metrics are grounded in the simple and intuitive idea that semantic change can be characterised by how well individual usages in one time period can be matched to semantically similar usages in another.

Across a wide range of experiments, we showed that these usage correspondence-based metrics are more robust to both representational quality and encoder choice. When paired with definition-based spaces, AMD improves in performance, demonstrating that interpretability and effectiveness need not be mutually exclusive. This opens up new possibilities for analysing semantic change in controlled and human-understandable spaces.

Overall, our findings show that AMD and SAMD should be considered alongside existing measures as a robust and flexible option for contextualised embedding-based semantic change detection. 

\section*{Limitations}
While our experiments show that APD and PRT degrade more sharply than AMD and SAMD under dimensionality reduction, and we offer hypothetical explanations for this trend, we do not provide a mechanistic account of the underlying causes. Understanding how different semantic change metrics interact with embedding geometry and variance structure remains an open question.

Although AMD and SAMD are more compatible with reduced and interpretable representation spaces, the metrics themselves yields scalar change scores (or two directional scalars), which are not directly interpretable on their own. 

Our robustness analysis is limited to different types of dimensionality reduction. We do not explore the behaviour of these metrics under other forms of embedding manipulation, such as noise injection, domain-adaptive fine-tuning, or alternative normalization and alignment procedures.

All definitions used in this work are generated using a single large language model (Gemini~2.5~Pro) with a fixed prompting strategy. While this ensures consistency across languages and experiments, variation in the quality, granularity, or coverage of the generated definitions may influence the resulting definition-based spaces and, in turn, the performance of AMD when paired with them. We do not evaluate the sensitivity of our results to alternative LLMs, prompting strategies, or numbers of generated definitions, and future work could explore how such factors affect the stability and interpretability of definition-based semantic change analysis.

While we evaluate AMD across multiple languages and encoders, our analysis is restricted to existing LSCD datasets with curated target words and gold change scores. The extent to which AMD facilitates lexical semantic change discovery, or interpretable analysis remains to be investigated.

\section{Acknowledgments}
We are grateful to Giacomo De Luca for his insightful feedback and suggestions.

This work has in part been funded by the research program Change is Key! supported by Riksbankens Jubileumsfond (under reference number M21-0021).



\appendix

\section{Number of Generated Definitions}
\begin{table}[h]
\centering
\begin{tabular}{lcc}
\hline
Language & \# Words & Defs / Word \\
\hline
English        & 37  & $11.00 \pm 5.36$ \\
German         & 48  & $3.75 \pm 1.70$ \\
Swedish        & 31  & $4.26 \pm 1.48$ \\
Latin          & 40  & $6.20 \pm 1.14$ \\
Spanish        & 100 & $6.84 \pm 4.14$ \\
Chinese        & 40  & $3.85 \pm 1.71$ \\
Norwegian\_1    & 40  & $4.75 \pm 1.88$ \\
Norwegian\_2    & 40  & $5.00 \pm 1.99$ \\
\hline
\end{tabular}
\caption{Number of target words and average number of generated definitions per word (mean $\pm$ standard deviation) for each language.}
\label{tab:num_defs}
\end{table}

\section{Definition Generation Prompts}
\begin{table*}[h]
\centering
\small
\setlength{\tabcolsep}{6pt}
\renewcommand{\arraystretch}{1.25}
\begin{tabularx}{\linewidth}{>{\raggedright\arraybackslash}p{2.2cm} X}
\hline
\textbf{Language} & \textbf{Prompt} \\
\hline

English &
Write all dictionary definitions of \texttt{\{WORD\}} in English.\newline
One sense per line. Output only the definitions. Do not add any text before or after the definitions. \\

German &
Schreibe alle Wörterbuchdefinitionen von \texttt{\{WORD\}} auf Deutsch.\newline
Eine Bedeutung pro Zeile. Gib nur die Definitionen aus. Füge keinen Text vor oder nach den Definitionen hinzu. \\

Swedish &
Skriv alla ordboksdefinitioner av \texttt{\{WORD\}} på svenska.\newline
En betydelse per rad. Skriv endast definitionerna. Lägg inte till någon text före eller efter definitionerna. \\

Latin &
Scribe omnes definitiones dictionarii verbi \texttt{\{WORD\}} Latine.\newline
Una significatio per lineam. Redde tantum definitiones. Ne quidquam addas ante aut post definitiones. \\

Spanish &
Escribe todas las definiciones de diccionario de \texttt{\{WORD\}} en español.\newline
Un significado por línea. Devuelve solo las definiciones. No añadas ningún texto antes ni después de las definiciones. \\

Chinese &
用中文写出“\texttt{\{WORD\}}”的所有词典释义。\newline
每行一个义项。只输出释义，不要在释义前后添加任何文字。 \\

Norwegian &
Skriv alle ordbokdefinisjoner av \texttt{\{WORD\}} på norsk.\newline
Én betydning per linje. Svar kun med definisjonene. Ikke legg til tekst før eller etter definisjonene. \\

\hline
\end{tabularx}
\caption{Language-specific prompts used for dictionary definition generation.}
\label{tab:definition_prompts}
\end{table*}

\section{Qualitative Analysis and Interpretability with AMD}

Beyond quantitative evaluation, AMD enables a more fine-grained and interpretable analysis of semantic change through its directional components.
Recall that the directional scores \(\mathrm{AMD}(A \rightarrow B)\) and \(\mathrm{AMD}(B \rightarrow A)\) capture asymmetric mismatches between periods.
High \(\mathrm{AMD}(A \rightarrow B)\) indicates usages in the earlier corpus that cannot be well matched in the later corpus, suggesting sense narrowing or disappearance, while high \(\mathrm{AMD}(B \rightarrow A)\) highlights usages emerging in the later corpus, corresponding to sense broadening or innovation.
Words exhibiting high values in both directions reflect overall shifts in meaning.

To illustrate this behaviour, we identify words with the largest imbalance between directional AMD scores (highest differences between them), highlighting cases of predominantly unidirectional change.
Across both full embedding space and definition-based space (produced with XL-LEXEME), we observe patterns of narrowing (e.g., \emph{plane}, \emph{head}) and broadening/emergence of new senses (e.g., \emph{record}, \emph{graft}).
While individual rankings vary slightly, the directional decomposition consistently surfaces words whose usage distributions change asymmetrically across time.

The definition-based space further enables a more explicit interpretation of these changes.
For a given target word, we apply linear discriminant analysis (LDA) to its usage embeddings projected into definition-based space, identifying the direction that best separates usages from the two time periods.
This yields a weighted combination of definition dimensions, where positive weights indicate stronger association with the later corpus and negative weights with the earlier corpus.
Inspecting the most discriminative dimensions provides a direct, human-interpretable signal as to which aspects of meaning are more or less represented between the two corpora (see \autoref{tab:qualitative_defs}).

For example, for \textit{plane}, LDA highlights a shift from tool-related meanings to air-travel-related usages, while for \textit{record}, the most discriminative dimensions correspond to a move from instrumental readings to official and documentary uses.

These examples illustrate how AMD, particularly when paired with interpretable lower-dimensional representations, supports targeted qualitative inspection of semantic change.
Rather than treating semantic shift as a single scalar quantity, this framework enables distinguishing between different change patterns—such as sense emergence, disappearance, or redistribution—and relating them to interpretable semantic dimensions, enabling fine-grained inspection of semantic change beyond what is possible with global distance metrics based on full embeddings.

\begin{table*}[t]
\centering
\small
\begin{tabular}{l p{8cm} p{6cm}}
\toprule
Word & Earlier-associated definition & Later-associated definition \\
\midrule
\textit{plane} &
A tool consisting of a block with a projecting blade, for shaping and smoothing a wooden surface. &
To travel in an airplane. \\
\textit{record} &
To indicate or show (a reading) on a measuring instrument. &
To state for the purpose of being set down in an official account. \\
\bottomrule
\end{tabular}
\caption{Most discriminative definition dimensions identified by LDA in definition-based space for selected words. Negative weights are associated with the earlier corpus and positive weights with the later corpus.}
\label{tab:qualitative_defs}
\end{table*}

\section{Hubness Analysis of Embedding Spaces}
\label{app:hubness}

Nearest-neighbour-based similarity measures in high-dimensional spaces are known to be affected by the \emph{hubness phenomenon}, whereby a small number of points appear disproportionately often as nearest neighbours of other points due to distance concentration and anisotropic embedding geometry. In the context of lexical semantic change detection, hubness may bias metrics such as Average Minimum Distance (AMD) if a small number of transitional or atypical historical usages act as hubs for many modern usages. For instance, a historical usage with unusually modern-like contextual properties could become the nearest neighbour for a large fraction of modern embeddings, leading AMD to underestimate broader distributional change.

To quantify hubness across embedding spaces, we compute three symmetric hubness statistics derived from nearest-neighbour occurrence distributions: (i) \textbf{dominant nearest-neighbour share}, the proportion of embeddings whose nearest neighbour is the single most frequent hub; (ii) \textbf{unused nearest-neighbour share}, the proportion of embeddings that are never selected as nearest neighbours (anti-hubs); and (iii) \textbf{average nearest-neighbour load}, the average number of embeddings assigned to each embedding that serves as a nearest neighbour at least once. Each measure is computed in both temporal directions and averaged to obtain symmetric scores.

\begin{table*}[h]
\centering
\begin{tabular}{lcccc}
\toprule
Metric & Full & Definition & PCA & Random \\
\midrule
Dominant NN share & .268 (.159) & .259 (.167) & .243 (.134) & .197 (.139) \\
Unused NN share & .616 (.069) & .578 (.100) & .583 (.090) & .497 (.049) \\
Average NN load & 2.882 (.618) & 3.175 (1.521) & 2.923 (.941) & 2.330 (.648) \\
\bottomrule
\end{tabular}
\caption{Mean (standard deviation) of hubness measures across spaces, aggregated over languages and encoders.}
\label{tab:hubness}
\end{table*}

Overall, hubness effects are moderate and comparable across spaces, and do not exhibit pathological collapse that would explain AMD’s performance gains. Dominant nearest-neighbour share remains around 2--27, indicating that only a minority of embeddings collapse onto a single hub, while unused nearest-neighbour share lies between 5 and 6, reflecting selective but non-pathological neighbourhood structure typical of high-dimensional embeddings. Importantly, the definition-projected space does not exhibit increased hub dominance relative to the original embedding space, and in fact slightly reduces both dominant and unused nearest-neighbour shares. The definition space shows a higher average nearest-neighbour load with greater variance, reflecting stronger local clustering around semantic prototypes induced by projection rather than pathological hub collapse.

To further assess robustness to hubness, we implement a symmetric variant of AMD based on greedy one-to-one matching between embeddings across corpora. Unlike standard AMD, which relies on independent nearest-neighbour assignments, this symmetric formulation constructs approximate optimal matchings and is therefore substantially less susceptible to hubness effects. Empirically, the symmetric matching AMD consistently outperforms APD and PRT and closely matches or improves upon standard AMD across spaces. This indicates that performance gains are not driven by hubness artefacts, but reflect improved semantic alignment between temporal corpora.

Overall, these results suggest that hubness does not account for the effectiveness of nearest-neighbour-based semantic change metrics in our setting, and that definition-based projection does not amplify hubness effects.

\end{document}